# DeepCFD: Efficient near-ground airfoil lift coefficient approximation with deep convolutional neural networks

Mohammad Amin Esabat[1] *, Saeed Jaamei[1], Fatemeh Asadi[2]

[1] Marine Department, Persian Gulf University, 7516913817, Bushehr, Iran
[2] Ocean Engineering Department, Texas A&M University, Galveston, TX 77554, USA
* Corresponding author, e-mail: Esabat@mehr.pgu.ac.ir

**Abstract**. Predicting and calculating the aerodynamic coefficients of airfoils near the ground with CFD software requires much time. However, the availability of data from CFD simulation results and the development of new neural network methods have made it possible to present the simulation results using methods like VGG, a CCN neural network method. In this article, lift-to-drag coefficients of airfoils near the ground surface are predicted with the help of a neural network. This prediction can only be realized by providing data for training and learning the code that contains information on the lift-to-drag ratio of the primary data and images related to the airfoil cross-section, which are converted into a matrix. One advantage of the VGG method over other methods is that its results are more accurate than those of other CNN methods.

## 1. INTRODUCTION

Fluid dynamics increasingly uses numerical techniques that allow us to calculate solutions to nonlinear equations that describe regimes and geometries relevant to engineering and industrial disciplines. Several numerical approaches have been developed theoretically and widely implemented on multiple parallel computers. In this paper, the VGG neural network is used to achieve the prediction of aerodynamic coefficients instead of heavy CFD calculations.

Researchers have proposed CNN models to predict flow and heat transfer characteristics around NACA sections and low Reynolds number airfoils. They also proposed a framework for designing low-Reynolds-number airfoils used in low-speed flights. To predict the aerodynamic coefficients of these airfoils, they have introduced a convolutional neural network (CNN) that uses XFOIL to generate data. They evaluated the CNN and found that it predicts the coefficients accurately. Later, they put the trained CNN together with an NSGA-II non-dominated sorting genetic algorithm to get the best results for multiple goals at a certain attack angle. They compared the optimization results with those obtained by directly invoking XFOIL and found that CNN significantly reduced the time required to generate Pareto-optimized fronts [1].

The use of machine learning in fluid mechanics has become so attractive that researchers, as part of their ongoing study, have created models using CNN and ED to predict the flow and heat transfer around NACA sections. The CNN predicted the aerodynamic coefficients and Nusselt number, while the ED model predicted the velocity, pressure, and thermal fields. The models were trained and tested using CFD simulations, with excellent agreement with real data. For the NACA cross-section design, predicted flow and thermal fields were used to interpret aerodynamic performance and heat transfer [2].

Researchers have conducted studies to predict static aerodynamic responses by combining the transform method and physics-based models. They have also used data-driven approaches with radial basis function neural networks (RBFNN) and convolutional neural networks (CNN) to predict unsteady aerodynamic responses and reconstruct flow fields, respectively [3]. In addition to applying machine learning to aerodynamic problems in airfoils. To solve this problem, flow control devices are implemented on airfoils. While computational fluid dynamics (CFD) simulations have been the leading method for analyzing these devices, using neural networks to predict flow characteristics is becoming increasingly popular. This study performed 158 CFD simulations on a DU91W(2)250 airfoil with rotating microtabs and Gurney flaps added to its trailing edge. The simulations were used to train and test convolutional neural networks (CNN) to predict the pressure and velocity fields, as well as predict the aerodynamic coefficient. The results showed that CNNs can accurately predict flow characteristics and aerodynamic coefficients while taking a lot less time to run than CFD simulations [4].

A dual neural network detection scheme is used, where the first neural network filters noisy images and the second CNN-based target detector extracts information from both radar data axes. The same CNN and a region-based convolutional network classify the detected targets as fixed-wing or rotary-wing aircraft. Experimental results show better performance compared to existing peak detection algorithms and CFAR on 1-D and 2-D data. A set of labelled Doppler range images from real-world air traffic control radar data has also been introduced [5].

Among the new machine learning methods. In this study, four methods for geometric input encoding were designed and compared. These methods produced accurate predictions with a mean absolute error of $1\times10^{-4}$. To improve accuracy near shock regions, wavelet transform, and gradient distribution loss were included. Pre-training the models on large-scale data sets and fine-tuning on small data sets improved their generalizability. Transfer learning reduced training time while maintaining comparable accuracy [6].

The results of the conducted studies show a minimal decrease in prediction accuracy with a significant decrease in optimization cost. Focusing on NACA airfoils, studies have successfully optimized the lift-to-drag ratio while comparing their findings with a high-order CFD solver. DeepONets exhibits little generalization error and enables accurate solutions for unseen shapes. This allows exploring objective functions beyond the lift-to-drag ratio [7].

To reduce the rework in the neural network, a real-time deep learning framework for detecting defective components in the aerospace industry is proposed. A convolutional neural network detects and classifies abnormal states during manufacturing. The proposed system can perform inspection and fault diagnosis, reducing the need for re-manufacturing. It also analyses the impact of production processes on real-time data. The results show a significant reduction in time delays and total cost [8].

Hui et al. and Duru et al. both used CNNs to predict the pressure distribution and flow field around airfoils. Hui et al. present a data-driven approach using a convolutional neural network (CNN) to predict pressure distribution over airfoils. It avoids the time-consuming task of computational fluid dynamics and the limitations of shape parameterization. By utilizing a signed distance function as a parametrization method, the model achieves accurate predictions of pressure coefficients with less than 2% mean square error in seconds. Duru et al. propose a machine-learning approach using an encoder-decoder convolutional neural network to estimate the pressure field around an airfoil. They trained and evaluated the network using computational fluid dynamics simulations of various airfoil shapes. To enhance shape learning, they calculated the pressure field using high-quality structured computational grids and provided a distance map of the airfoil shape. Their model achieved 88% accuracy for unseen airfoil shapes and showed promising results in capturing the overall flow pattern. It also significantly sped up the process compared to traditional CFD simulations [9][10].

Liu et al. and Sekar et al. both used CNNs to predict flow fields over airfoils. Yunzhu's model was better at predicting flow fields in real time than traditional machine learning methods. Yunzhu et al. discussed a supervised learning method using deep convolutional neural networks (CNN) for reconstructing physical fields in a heat transfer problem. The CNN predicts fields based on a limited amount of measurable information, allowing for easy inference of heat transfer characteristics. The method can serve as a surrogate model for computational fluid dynamics (CFD) when accurate structure or work condition parameters are available. It can also reconstruct full fields from local data with multiple measuring points as inputs. The CNN model proves to be a high-fidelity field predictor for flow heat transfer, accurately reconstructing temperature, velocity, and pressure fields, as well as extracting heat transfer process characteristics. Various perspectives verify the reconstruction performance and stability, demonstrating the significantly faster performance of the CNN model over the CFD solver. Overall, this approach provides an efficient analysis tool for heat transfer research with acceptable accuracy [11][12]. Lou et al. propose a method for aerodynamic optimization of airfoils using a combination of deep learning and reinforcement learning. A deep neural network is used to predict the lift-drag ratio, and a double deep Q-network algorithm is designed to train the optimization policy. The results show that the method can improve the lift-drag ratio to 71.46% and adapt to diverse computational conditions. Overall, this approach allows for rapid prediction of aerodynamic parameters and intelligent optimization [13]. In addition, Ye et al.'s study compared CNN output data with computational fluid dynamics (CFD) results. This showed that CNN was very good at predicting pressure coefficients. This comparison provides empirical evidence supporting the accuracy of CNN-based predictions in aerodynamics [14]. Additionally, Bhatnagar et al. highlighted the enhanced predictive capabilities of CNN due to specific convolution operations, parameter sharing, and robustness to noise, further reinforcing the effectiveness of CNN in aerodynamic predictions [15].

Trailing edge flaps (TEFs) are high-lift devices that alter lift and drag coefficients on airfoils. Rodriguez-Eguia et al. performed numerous 2D simulations to measure these changes and analyze them for specific parameters. Three airfoils (NACA 0012, NACA 64(3)-618, and S810) were studied. The results validate expected aerodynamic effects, demonstrating the impact of parameters on lift and drag coefficients. Flap deployment in different zones affects the lift-to-drag ratio (CL/CD), while larger flap lengths adjust the ratio's values. Additionally, an artificial neural network (ANN) was created to predict aerodynamic forces based on the research findings [16].

Xie et al. suggest using a knowledge-embedded meta-learning model to guess the lift coefficients of an airfoil. The model uses a primary network to find the lift-angle relationship and a hyper network to find information about the geometry of the airfoil. The model shows strong generalization and accuracy and can analyze the influence of airfoil geometry on characteristics [17].

According to Olayemi et al., airfoil selection is critical in aircraft design, influencing wing lift and fuselage drag. Traditional methods for obtaining aerodynamic coefficients involve costly CFD simulations. These models accurately predicted airfoil lift coefficients, offering an alternative to time-consuming simulations [18].

The present study utilizes a convolutional neural network (CNN) to solve a regression problem, specifically predicting aerodynamic coefficients such as lift coefficients from CFD simulation data. The model employed is based on a modified version of the VGGNet architecture, which has been adapted for regression tasks. The model uses the means squared error (MSE) loss function, a standard choice for regression problems, to minimize the error between predicted and true aerodynamic coefficients. The architecture comprises two convolutional layers, followed by two fully connected (FC) layers. The model is trained and evaluated using supervised data derived from CFD simulation results.

## 2. METHODS

The approach uses a convolutional neural network (CNN) to predict aerodynamic coefficients from CFD simulation data. CNNs are particularly well-suited for regression tasks involving high-dimensional data, such as image-like representations of airflow around airfoils. The model is based on a modified VGGNet architecture, adapted specifically for the task of predicting aerodynamic coefficients. In this architecture, the input data representing airfoil geometries or flow field results undergoes feature extraction through convolutional layers before being processed by fully connected layers to output the predicted aerodynamic coefficients. The network is trained using supervised data, with mean squared error (MSE) used as the loss function to optimize the model's predictions.

## 2.1 CNN (Convolutional neural networks)

Convolutional neural networks (CNNs) are well-suited for pattern recognition in multidimensional data, such as images or matrices. CNNs typically consist of convolutional layers that use kernels (filters) to extract hierarchical features from the input data. These are followed by one or more fully connected layers to integrate the extracted features and produce the output. CNNs are widely used for tasks such as image classification, object detection, and regression tasks, making them a natural choice for predicting aerodynamic coefficients from image-like representations of airfoil geometries or simulation results.

In the modified VGGNet architecture used in this study, the input data is first processed through two convolutional layers. These layers apply a series of filters to detect low- and high-level features of the input image or matrix, such as edges and textures. Each convolutional layer is followed by a ReLU activation function, which introduces non-linearity and allows the network to model complex, non-linear relationships between the input data and the predicted aerodynamic coefficients. The output of the convolutional layers is flattened and passed through two fully connected (FC) layers. These FC layers are responsible for combining the learned features from the convolutional layers and producing the final output the predicted aerodynamic coefficient values.

Convolutional neural networks (CNN) use a specific architecture to recognize patterns in multidimensional data such as images [19]. These networks consist of convolutional layers that use filters or kernels to extract different image features. CNNs typically include fully connected layers to combine features from convolutional layers. CNNs are widely used in tasks such as object recognition in images, face recognition, machine image translation, and more. Figure 1 shows an example of the CNN network model, which has four convolutional layers and, at first, receives the data input in the form of a matrix, and then the 1st and 2nd Fully Connected layers provide the desired output. Eq. (1) represents the mathematical formulation of the convolutional layer operation:

$$Y[i,j] = \sigma\left(\sum_{m=0}^{M-1}\sum_{m=0}^{N-1} X[i+m, j+n]\right) \cdot W[m,n] + b \qquad (1)$$

Here, $X$ is the input image or matrix, $W$ represents the convolutional kernel (filter layer, $b$ is the bias term, Y, and $\sigma$ is the chosen activation function (ReLU). For fully connected layers, the mathematical operation is formulated as follows:

$$Y = \sigma(W \cdot X + b) \qquad (2)$$

$X$ is the flattened input vector from previous layer, $W$ is the weight matrix, $b$ is the bias term. Again, $\sigma$ is the activation function that is applied after the linear transformation[20].

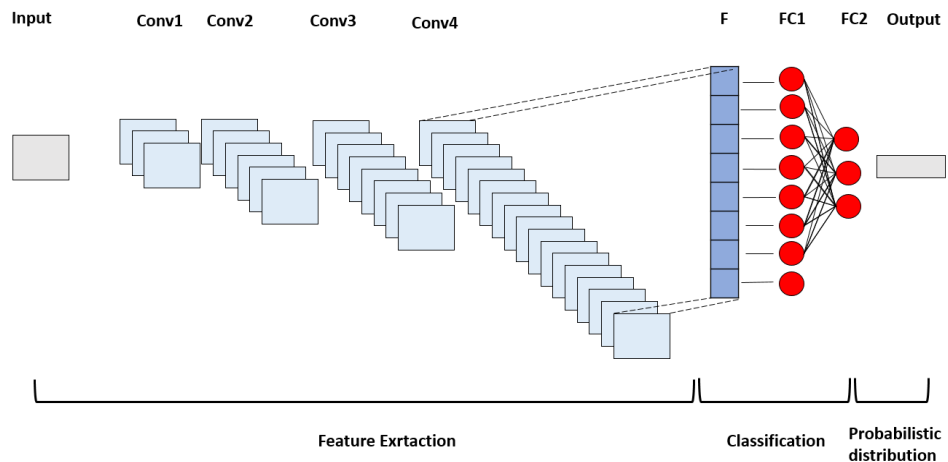

**Fig. 1.** Typical convolutional neural network.

### 2.1.1 Kernel (Convolutional Kernel)

According to Figure 2, the kernel is a small matrix (or filter) used in convolutional operations within convolutional layers. The kernel slides across the input data, computing dot products between the kernel and the receptive field of the input to extract specific features. Kernels are crucial for detecting patterns such as edges, textures, or shapes in images. By applying multiple kernels, the network can learn hierarchical feature representations, starting from simple patterns in early layers to more complex ones in deeper layers.

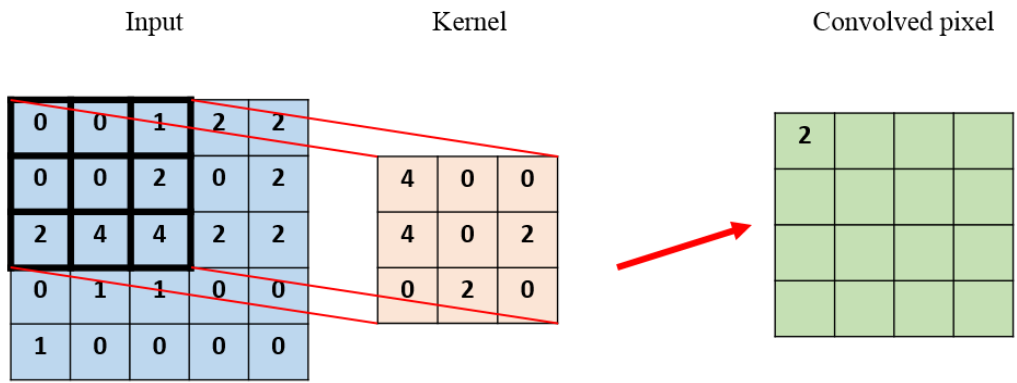

**Fig. 2.** Typical convolution operation in the CNN

### 2.1.2 Dimensionality Reduction

Dimensionality reduction is reducing the number of features or data points in a dataset. In the context of CNNs, this often refers to reducing the input data's spatial dimensions (width and height) of the input data, particularly through techniques like pooling. Dimensionality reduction helps manage computational complexity, reduces overfitting, and extract essential information from input data.

### 2.1.3 Pooling Layer

As shown in Figure 3, the pooling layer is used to reduce the spatial dimensions of the input data. Common pooling operations include max pooling, where the maximum value from each region is retained, and average pooling, where the average value is kept. Pooling helps reduce the computational load and makes the network less sensitive to small translations or distortions in the input data. This also encourages the model to focus on the most important features. The mathematical relationship for the pooling operation is represented by Eq. (3):

$$Y[i,j] = \max\left(X[2i,2j], X[2i,2j+1], X[2i+1,2j], X[2i+1,2j+1]\right) \tag{3}$$

Here, $X$ is the input to the pooling layer, and $Y$ is the layer output [21].

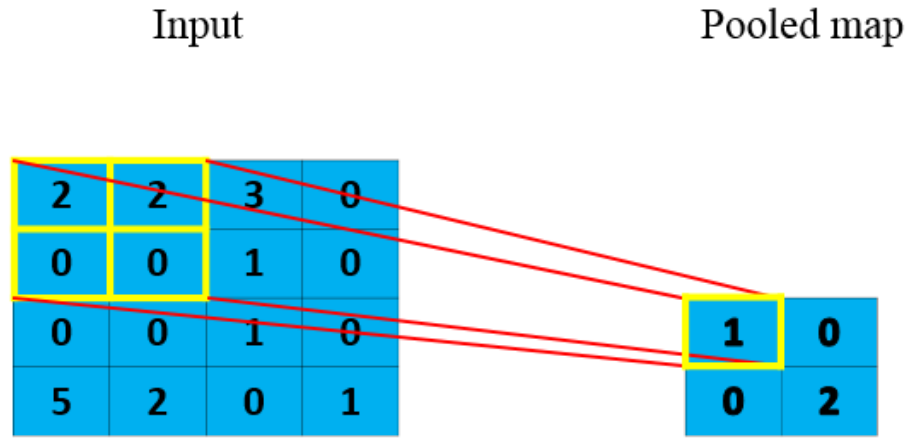

**Fig. 3.** Typical pooling operation in the CNN

### 2.1.4 ReLU (Rectified Linear Unit)

ReLU is an activation function applied element-wise to the output of a neuron in a neural network. It replaces all negative values with zero and leaves positive values unchanged. This function introduces non-linearity to the model, enabling it to learn complex patterns. ReLU also helps the network converge faster during training by avoiding issues like vanishing gradients that are encountered with other activation functions, such as sigmoid or tanh [22].

$$\operatorname{Re}LU(x) = \max(0, x) \tag{4}$$

where x is the input or the ReLU function.
These components collectively contribute to the effectiveness of convolutional neural networks in image recognition and other computer vision tasks [23].

### 2.1.5 Governing Equations: Navier-Stokes Equations

The presence of the ground significantly influences the aerodynamic performance of an airfoil, particularly through the ground effect, which alters lift and drag characteristics near the ground surface. This effect is most notable during take-off and landing. The governing equations for fluid dynamics are derived from the Navier-Stokes equations. Eq. (5) represents the conservation of mass

(continuity equation), Eq. (6) corresponds to the momentum equation, and Eq. (7) describes the conservation of energy.

$$\frac{\partial \rho}{\partial t} + \nabla \cdot (pu) = 0 \tag{5}$$

$$\frac{\partial (\rho u)}{\partial t} + \nabla \cdot (pu \otimes u) = -\nabla p + \nabla \cdot (\mu \nabla u) + f \tag{6}$$

$$\frac{\partial (\rho E)}{\partial t} + \nabla \cdot (puH) = \nabla \cdot (\mu \nabla u) + \nabla \cdot (\kappa \nabla T) + \dot{q} \tag{7}$$

where $\rho$ is the density, $u$ is the velocity vector, $p$ is pressure, $\mu$ is dynamic viscosity, $\kappa$ is thermal conductivity, $f$ represents external forces, $H$ is the specific enthalpy, $E$ is the total energy density, $T$ is temperature, and $\dot{q}$ is the dissipation term[24].

### 2.1.6 Numerical Methods: Finite Volume Approach

In this study, the finite volume method (FVM) was employed to numerically solve the Navier-Stokes equations and simulate the airfoil-ground interaction. The computational domain is discretized into finite control volumes, and the governing equations are solved on each of these volumes. FVM is a widely used numerical technique for solving partial differential equations (PDEs), particularly in fluid dynamics, where it conserves physical quantities such as mass, momentum, and energy. A time-stepping algorithm is used to advance the solution over time, where the integral form of the conservation equations is discretized over the control volumes.

For the scalar quantity $\varphi$, the finite volume formula can be written in the following general form as Eq. (11):

The integral form of the mass conservation equation can be expressed as:

$$\frac{d}{dt} \int_v \rho dV + \int_s \rho \mathbf{u} \cdot \mathbf{n} dA = 0 \tag{8}$$

Where:
- ρ is the fluid density.
- V is the control volume.
- S is the boundary surface of the control volume.
- **u** is the velocity vector.
- **n** is the outward normal vector to the surface.

The integral form of the momentum equation can be expressed as:

$$\frac{d}{dt} \int_v \rho \mathbf{u} dV + \int_s \rho \mathbf{u}(\mathbf{u} \cdot \mathbf{n}) dA = -\int_s p\mathbf{n} dA + \int_s \tau \cdot \mathbf{n} dA + \int_v \mathbf{f} dV \tag{9}$$

Where:
- **u** is the velocity vector.
- p is the pressure.
- τ is the viscous stress tensor.

- **f** is the body force per unit volume.

The integral form of the energy conservation equation is written as:

$$\frac{d}{dt}\int_v \rho e dV + \int_s \rho e \mathbf{u} \cdot \mathbf{n} dA = \int_s \rho e \mathbf{u} \cdot \mathbf{n} dA + \int_v Q dV \tag{10}$$

Where:
- e is the specific internal energy.
- q is the heat flux.
- Q represents any volumetric heat sources.

$$\frac{\partial(\rho V \varphi)}{\partial t} + \sum_f F_f = 0 \tag{11}$$

where $\rho$ is the density, $V$ is the volume of the control volume, $\varphi$, is the scalar quantity (e.g., concentration, temperature), $F_f$ represents the flux through the forces of the control volume, and the sum is taken over all forces $f$ of the control volume [25].

*2.1.7 Computational Considerations: Grid Resolution and Turbulence Modelling*

The accuracy of the simulation is highly dependent on both grid resolution and turbulence modeling. A fine grid resolution near the airfoil surface and the ground is employed to accurately capture boundary layer effects, which are crucial for predicting aerodynamic forces. Additionally, turbulence modeling is essential to simulate the turbulent flow conditions around the airfoil. In this study, a Large Eddy Simulation (LES) model is implemented to resolve large-scale turbulent eddies and model the subgrid-scale turbulence.

The filtered Navier-Stokes equations for large eddy simulation (LES) can be expressed as:

$$\frac{\partial \tilde{u}}{\partial t} + (\tilde{u} \cdot \nabla)\tilde{u} = -\frac{1}{\rho}\nabla \tilde{\rho} + \nu \cdot \nabla^2 \tilde{u} - \nabla \cdot \tau + f \tag{12}$$

where $\tilde{u}$ represents the filtered velocity field, $\tilde{\rho}$ is the filtered pressure, $\rho$ is the fluid density, $\nu$ is the kinematic viscosity, $\tau$ denotes the subgrid-scale stress tensor, and $f$ represents external forces [26].

**2.2 VGG**

VGG is a deep convolutional neural network architecture developed by the Visual Geometry Group at the University of Oxford. It is well-known for its deep structure and its use of small convolutional kernels, particularly 3x3 filters, which allow the network to focus on local patterns while maintaining a manageable number of parameters. VGG's architecture, which can go up to 19 layers (e.g., VGG-16 or VGG-19), has been highly influential in the field of image classification and other computer vision tasks.

In Figure 4, the overall process is shown as a flowchart. In this case, the inputs are received in binary form and redefined as a matrix, while the neural network is trained and tested, and finally the coefficients are predicted.

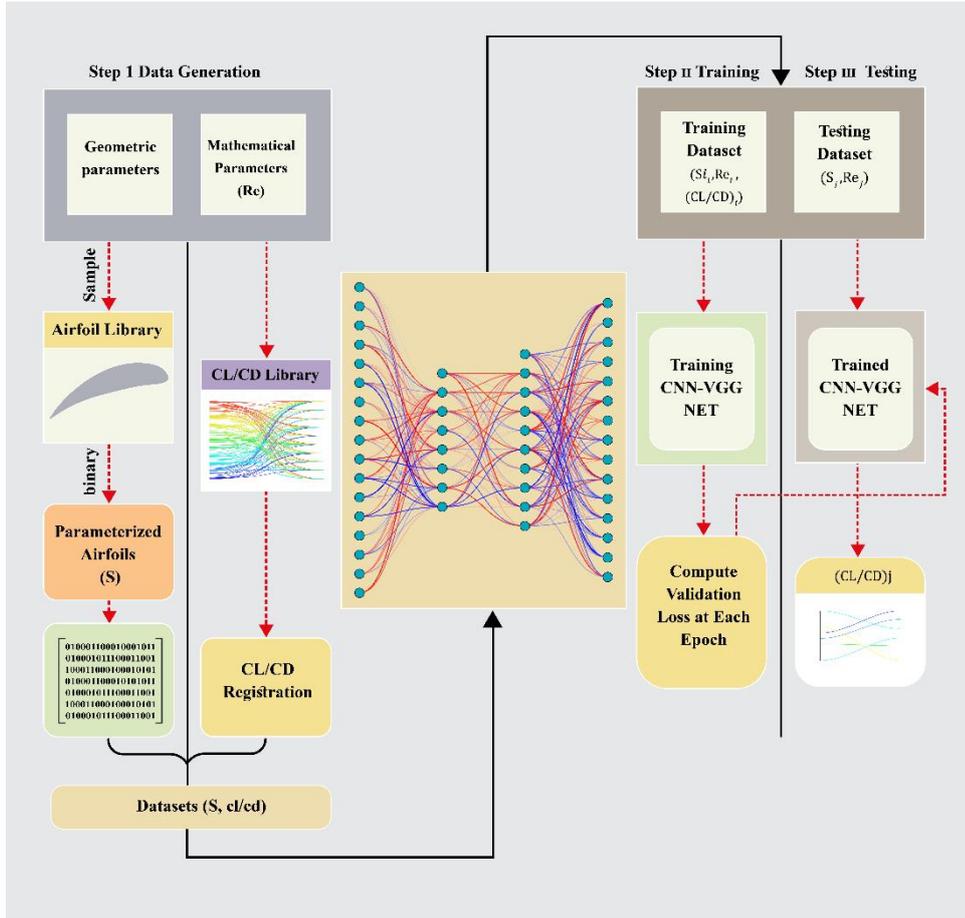

**Fig. 4.** Overall architecture of CNN-VGG framework.

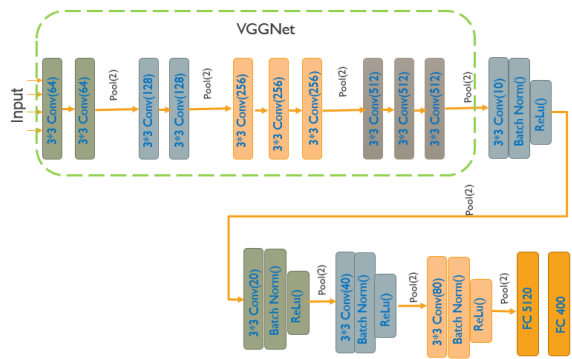

**Fig. 4.** VGG neural network model

### 2.2.1 Key aspects of VGG

The pattern in Figure 4 illustrates the general architecture of the VGG method used in this study. Key aspects of the VGG architecture include:
1. Deep Structure: VGG has a deep architecture, often consisting of up to 19 layers, which enables the model to learn complex, hierarchical feature representations.
2. Small Kernels: VGG employs 3x3 convolutional kernels in all convolutional layers, which allows the model to focus on local features while keeping the number of parameters relatively low.
3. Downsampling: Max pooling layers with a 2x2 filter and stride 2 are used for dimensionality reduction, progressively reducing the spatial dimensions while retaining essential features.

4. Batch Normalization: In some versions of VGG (such as VGG-19), batch normalization layers are included to stabilize and accelerate training by normalizing the output of the previous layer.

VGG has laid the foundation for many subsequent advancements in neural network architectures, particularly in the domain of image recognition. Variants such as "VGG-16" or "VGG-19" differ in the number of layers, with VGG-16 having 16 layers and VGG-19 having 19 layers [27].

VGGs feature extraction method by convolution operation also makes it detect local features better. It also has fewer parameters to learn because of parameters sharing mechanism. In this project, we choose VGG as our model The VGG architecture is composed of two parts: Encoder, which extracts features, and Decoder, which learns the features and outputs regression results. The Encoder contains 4 convolutional layers. In each convolutional layer, batch normalization, an optimization technique, is applied to shift and scale the data from last layer.It has been shown that by batch normalizing neural networks can be trained faster and have higher accuracy. Then features are extracted by applying kernels filtering the image and obtained feature maps.

## 3. DATA PROCESSING

In this study, a data processing and image generation approach is employed to investigate the relationship between the lift-to-drag ratio and airfoil shape using a neural network. The first step in this process involves preparing a sufficiently diverse and quantitatively adequate dataset. To achieve this, the UIUC Airfoil Database, which contains the coordinates of 2200 airfoils, is utilized as the primary source of input data.

The raw coordinate pairs for each airfoil are first transformed into grayscale images (128x128 pixels) to address the challenge of non-uniform data sizes. This transformation allows the model to work with consistent input dimensions. The contour plots of the airfoils are generated, followed by morphological operations to enhance the features. For each airfoil, 25 images are produced at different angles of attack, allowing the neural network to learn the variation in aerodynamic characteristics with respect to the angle of attack.

The generated images are then flattened and binarized, forming a data matrix of dimensions N×224848, where N represents the number of airfoils in the dataset. This matrix serves as the input for training the neural network.

For validation, the study uses results from well-established aerodynamic analysis software: XFLR5, XFOIL, and CFD simulations. Additionally, original data from NASA's airfoil database is used to ensure the accuracy and reliability of the model's predictions.

## 4. PROBLEM SETUP

This study introduces a convolutional neural network (VGG) for predicting the lift-to-drag ratio of airfoils. To evaluate the efficiency and accuracy of the model, several performance metrics are employed, including the mean squared error (MSE), confusion matrix, and computation time.

The dataset used consists of 2200 airfoil samples, from which 1000 samples were randomly selected for training and evaluation. The dataset is split into three subsets: 70% for training, 20% for validation, and 10% for testing. The training and validation subsets are used to adjust the model's parameters, while the testing subset is used to evaluate its generalization performance.

The angle of attack for each airfoil varies from 0° to 20° in increments of 0.25°. For each airfoil sample, 100 images are generated at different angles of attack, yielding a total of 20,578 images from the 1000 airfoil samples. These images are converted into matrices of the same size, which also serve as the labels for the prediction task.

The model's hyperparameters were set as follows:

- Batch size: 50
- Learning rate: 0.00005
- Epochs: 100

The validation dataset plays a crucial role in adjusting model parameters, as it is used to monitor and minimize the discrepancy between training and validation performance over time.

## 5. RESULTS

In this section, the performance evaluation of the VGG model is presented. After processing the data, the model is trained, and its performance is assessed using error curves, a confusion matrix, and metrics such as accuracy and efficiency.

### 5.1 Error Curves on the Test-Set

Figure 6 illustrates the performance of our VGG model over 100 epochs. Both the training and validation error curves converge after approximately 50 epochs, indicating that the model has effectively learned the underlying patterns in the data. The training loss stabilizes at 0.36484, while the validation loss reaches 0.06415 at the end of the training.

Despite some fluctuations in the number of adjustable parameters and learning rate settings, the training and validation loss curves exhibit similar trajectories, suggesting that the model's learning process is stable. These error curves demonstrate that the VGG model achieves a good fit on both the training and validation sets, reaching the theoretical limit of achievable accuracy for predicting the lift-to-dra coefficient.

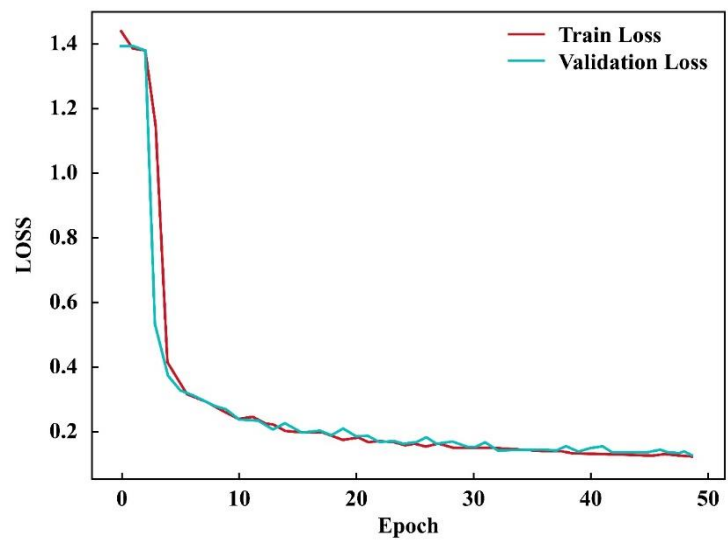

**Fig. 5.** Train vs. Valid Loss

### 5.2 Confusion Matrix

Figure 7 illustrates the deviation between the ground truth and the predicted values from the test dataset. The x-axis represents the predicted values, while the y-axis represents the actual ground truth. If the VGG model's predictions are accurate, the points along the diagonal will appear red, indicating that the predicted values closely match the true values. A noticeable improvement can be seen from epoch 5 to epoch 300, demonstrating the effectiveness and progress of the VGG model in making accurate predictions.

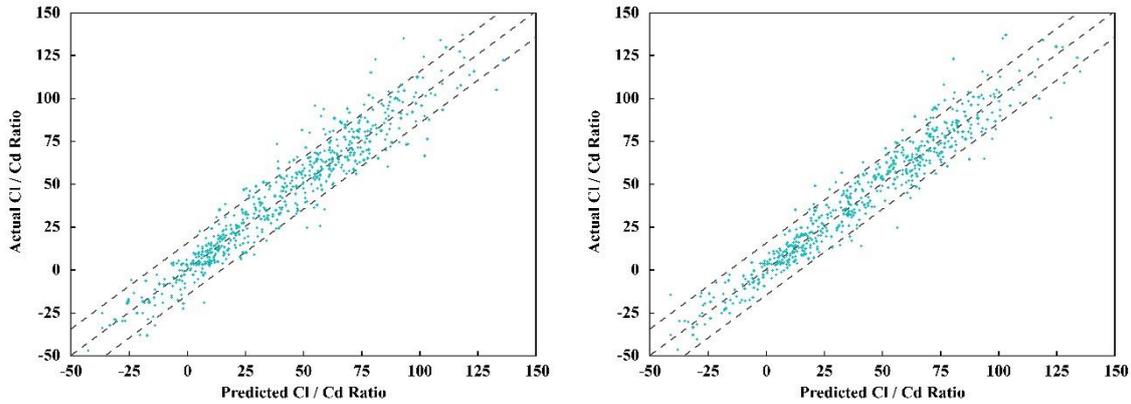

**Fig. 6.** Confusion matrices with different training epochs (left with 50 training epochs and right with 200 training epochs)

### 5.3 Prediction and Ground Truth

Figure 8 compares the predictions generated by the VGG model with the actual ground truth data. The orange lines represent the ground truth lift-to-drag ratios (CL/CD) for each airfoil sample at a specific angle, as calculated through computational fluid dynamics (CFD). In contrast, the blue lines show the predictions made by the VGG model. The x-axis represents the airfoil sample serial numbers, while the y-axis displays the lift-to-drag ratio. In addition, it provides a more detailed comparison for a single airfoil. Notably, both plots exhibit a significant overlap, demonstrating the high accuracy of the model's predictions, which closely align with the ground truth.

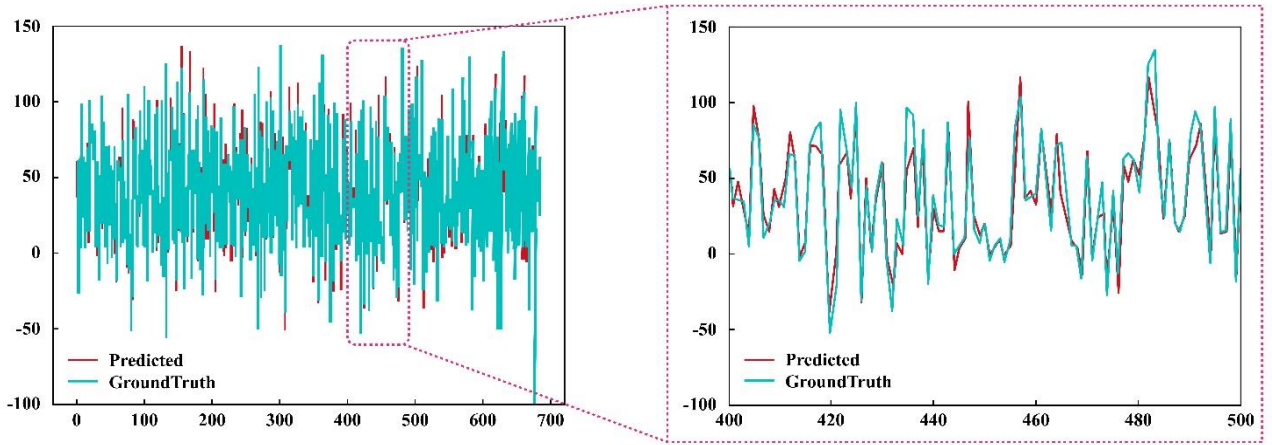

**Fig.7.** Test and Predicted Cl/Cd Ratio and Test and Predicted Cl/Cd Ratio (Zoon In)

### 5.4 Efficiency

In contrast to the computational fluid dynamics (CFD) method, which requires multiple iterations and extensive calculations to achieve converged results, neural networks offer a significantly faster alternative. The CFD solver processes each sample individually, performing numerous iterations to reach accurate outcomes, which can be time-consuming. In contrast, once trained, neural networks can produce accurate results much more efficiently, without the need for repeated testing. Overall, the use of neural networks dramatically reduces computational time and resources, offering a more efficient approach for predicting aerodynamic coefficients.

# 6. CONCLUSION

Based on the results presented, the following conclusions can be drawn:

- The use of a Convolutional Neural Network (CNN), specifically based on the VGG architecture, has demonstrated the capability to maintain high accuracy in predicting aerodynamic coefficients, particularly the lift-to-drag ratio. The model's performance improves with an increasing number of training epochs, highlighting the effectiveness of deeper architectures in learning complex relationships.
- With its high accuracy, the CNN model can be extended to optimize airfoil geometry, predict flow behaviors around hydrofoils, and simulate fluid dynamics in different environments such as water and air. This makes the method versatile for a wide range of applications in aerodynamics and hydrodynamics.
- Neural networks have the potential to revolutionize industries by simulating complex fluid dynamics problems governed by the Navier-Stokes equations. By reducing the need for time-consuming CFD simulations, they provide a cost-effective alternative for both academic research and industrial applications, improving efficiency and reducing computational time.
- The availability of large, high-quality datasets plays a crucial role in the success of machine learning models in predicting outcomes for various fluid-related problems. As more data becomes available, the accuracy of such models will continue to improve, and the scope of their applications will expand.
- Methods like transfer learning also offer significant promise in overcoming the challenges posed by limited datasets. By leveraging pre-trained models on large-scale data, transfer learning enables the efficient application of machine learning techniques even with smaller datasets, opening new opportunities for predictive modeling in fluid dynamics.

In conclusion, the integration of machine learning techniques, particularly CNNs, into the field of fluid dynamics offers promising advancements in both simulation and optimization, significantly reducing computational costs and time. However, further studies are needed to refine the models, explore their potential in other fluid-related areas, and assess their generalization capabilities for real-world applications.

**Ethical Approval**

This research did not contain any studies involving animal or human participants, nor did it take place on any private or protected areas. No specific permissions were required for corresponding locations.

**Competing Interests**

We have no competing interests.

**Funding**

There is no funding applicable to this article.

**Availability of Data and Materials**

Datasets generated and/or analyzed during the current study are available at sciencedb, [10.57760/sciencedb.13476]. This access will be open to the public from the date of 01/08/2025. Also, datasets generated and/or analyzed during the current study are provided to the corresponding author upon request.